%% file: main.tex
\documentclass[10pt,twocolumn,letterpaper]{article}

\usepackage{cvpr}              
\usepackage{multirow}
\usepackage[table]{xcolor}
\usepackage{algorithm}
\usepackage{algorithmic}
\usepackage{amsmath}
\usepackage{bbding}
\usepackage{amssymb}
\usepackage{cases}
\usepackage{listings}

\PassOptionsToPackage{prologue, dvipsnames}{xcolor}
\input{preamble}

\definecolor{cvprblue}{rgb}{0.21,0.49,0.74}
\usepackage[pagebackref,breaklinks,colorlinks,citecolor=cvprblue]{hyperref}

\def\paperID{12381} 
\def\confName{CVPR}
\def\confYear{2024}

\title{BlindDiff: Empowering Degradation Modelling in Diffusion Models for Blind Image Super-Resolution}

\author{
    Feng Li$^{1}$ \quad Yixuan Wu$^{2}$ \quad Zichao Liang$^{2}$ \quad Runmin Cong$^{3}$ \quad Huihui Bai$^{2, \dagger}$ \quad Yao Zhao$^{2}$ \quad Meng Wang$^{1}$\\
    $^{1}$Hefei University of Technology $^{2}$Beijing Jiaotong University $^{3}$Shandong University\\
}

\begin{document}
\maketitle
\input{sec/0_abstract}    
\input{sec/1_intro}
\input{sec/2_formatting}
\input{sec/3_preliminary}
\input{sec/4_method}
\input{sec/5_experiments}
\input{sec/6_conclusion}
{
    \small
    \bibliographystyle{ieeenat_fullname}
    \bibliography{main}
}
\end{document}

%% file: preamble.tex
\usepackage[dvipsnames]{xcolor}


%% file: sec/0_abstract.tex
\begin{abstract}
Diffusion models (DM) have achieved remarkable promise in image super-resolution (SR). However, most of them are tailored to solving non-blind inverse problems with fixed known degradation settings, limiting their adaptability to real-world applications that involve complex unknown degradations. In this work, we propose BlindDiff, a DM-based blind SR method to tackle the blind degradation settings in SISR. BlindDiff seamlessly integrates the MAP-based optimization into DMs, which constructs a joint distribution of the low-resolution (LR) observation, high-resolution (HR) data, and degradation kernels for the data and kernel priors, and solves the blind SR problem by unfolding MAP approach along with the reverse process. Unlike most DMs, BlindDiff firstly presents a modulated conditional transformer (MCFormer) that is pre-trained with noise and kernel constraints, further serving as a posterior sampler to provide both priors simultaneously. Then, we plug a simple yet effective kernel-aware gradient term between adjacent sampling iterations that guides the diffusion model to learn degradation consistency knowledge. This also enables to joint refine the degradation model as well as HR images by observing the previous denoised sample. With the MAP-based reverse diffusion process, we show that BlindDiff advocates alternate optimization for blur kernel estimation and HR image restoration in a mutual reinforcing manner. Experiments on both synthetic and real-world datasets show that BlindDiff achieves the state-of-the-art performance with significant model complexity reduction compared to recent DM-based methods. Code will be available at \url{https://github.com/lifengcs/BlindDiff}
\end{abstract}

%% file: sec/1_intro.tex
\section{Introduction}
\label{sec:intro}
Single image super-resolution (SISR) is a long-standing problem in computer vision and image restoration, which has attracted significant attention due to its ill-posed nature and broad practical applications. SISR refers to generating a high-resolution (HR) image $\hat{\mathbf{x}}$ from a low-resolution (LR) observation $\mathbf{y}$ given through a degradation model 
\begin{equation}
\mathbf{y}=(\mathbf{k}\otimes\mathbf{x})\downarrow_s+\mathbf{n}
\label{eq0}
\end{equation}
where $\mathbf{k}$ denotes the blur kernel convolved with $\mathbf{x}$, followed by a downsampling operation with scale factor $s$. $\mathbf{n}$ is usually assumed as the additive white Gaussian noise (AWGN). Over the past years, methods for SISR have been dominated by deep learning approaches~\cite{srcnn,rcan,esrgan,swinir,fdiwn,cpwsnn,hat}. Most of them mainly work on pre-defined degradations (\emph{e.g.} bicubic downsampling), which train powerful deep convolutional networks with end-to-end supervision on synthesized numerous LR-HR pairs. Despite their success, they would deteriorate in performance when applied to other unseen degradations. This is particularly evident in real world where the degradations are complicated and unavailable. 
\begin{figure}
  \centering
   \subfloat[FID \emph{v.s.} Model parameter.]{
    \label{fig:a}
    \includegraphics[width=0.455\linewidth]{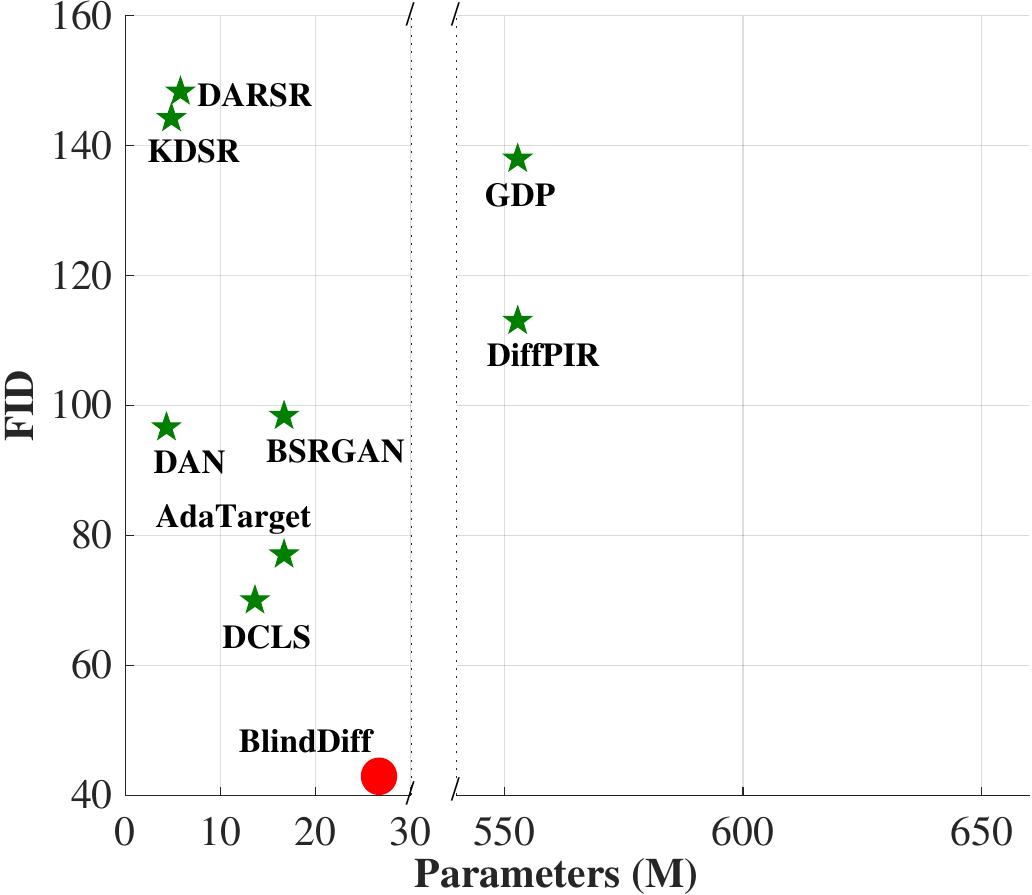}
}
   \subfloat[LPIPS \emph{v.s.} Multi-Adds (G).]{
   \label{fig:b}
    \includegraphics[width=0.49\linewidth]{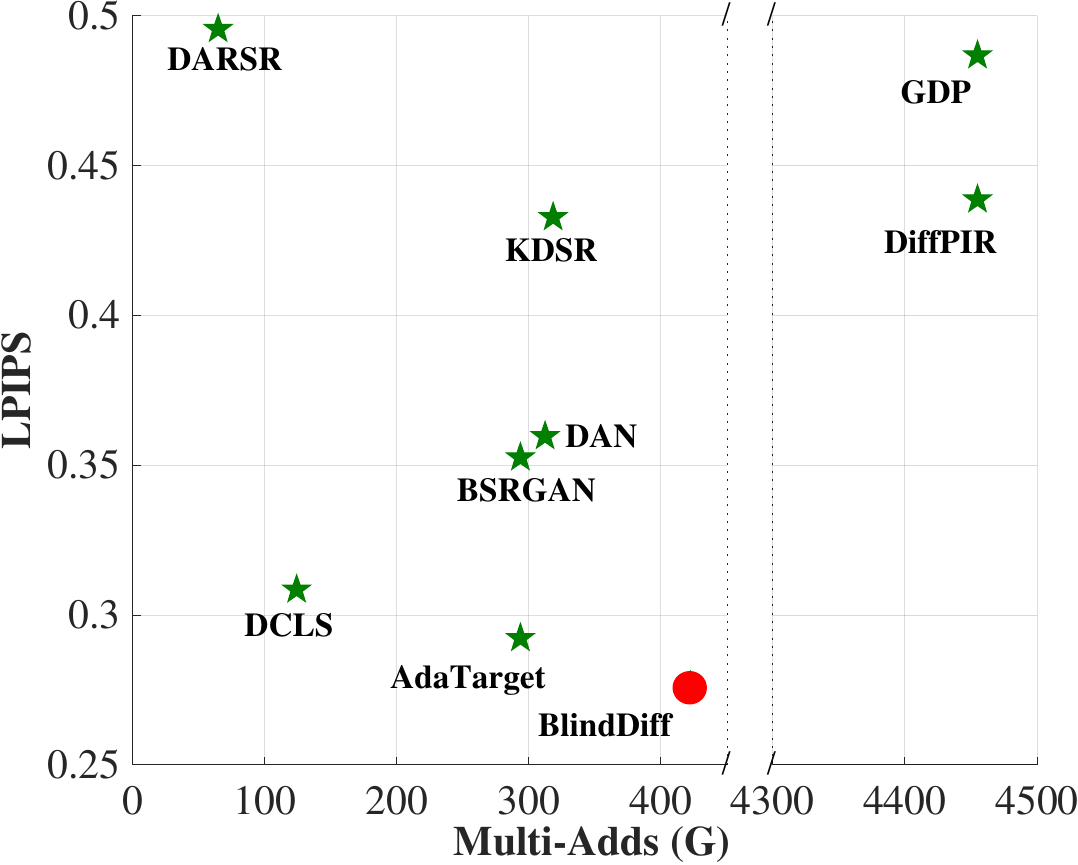}
}
  \caption{Our BlindDiff achieves the state-of-the-art performance for $4\times$ blind SR (Table~\ref{tab2}) which being more computational efficient than existing DM-based methods. The Multi-Adds are computed based on the LR size of $128\times 128$.}
\end{figure}

To overcome this limitation, recent researchers pay increasing attention to blind image super-resolution (SR)~\cite{kernelgan,ikc,dasr,dssr}, where their key steps lie in degradation estimation (mostly blur kernel $\mathbf{k}$) and degradation-aware SR reconstruction. One option~\cite{kernelgan,manet} is to estimate the underlying $\mathbf{k}$ from LR input $\mathbf{y}$ and apply that to non-blind SR models~\cite{zssr,usrnet} for HR image generation. Another approach~\cite{ikc,dan,kxnet} is to unify kernel estimation and SR reconstruction in a single end-to-end framework. However, their results still suffer from noticeable artifacts and low perceptual quality in dealing with complex degradations. 

Very recently, the emerging diffusion models (DM)~\cite{dm,ddpm} have been demonstrated as a powerful generative model for high-fidelity image generation~\cite{dalle2,ldm,ddrm,frido}. Currently, there are two main paradigms of DM-based methods for SISR. One is to train the DM from scratch conditioned on LR images~\cite{srdiff,sr3}. The other more popular is to design problem-agnostic models~\cite{ddrm,mcg,ddnm,dps}, which employs off-the-shelf pre-trained DMs as the generative prior for general-purpose restoration. Nevertheless, it is important to note that both approaches typically focus on non-blind inverse problems, meaning that the image degradation is assumed to be known and fixed. This is impractical in real-world applications, thereby severely restricting their generalization ability.

In this work, we explore the potential of DMs for blind SR and propose BlindDiff, a novel diffusion super-resolver that is flexible and robust to various degradations. This method is inspired by recent denoising diffusion probabilistic models (DDPM)~\cite{ddpm,dps,score} for non-blind inverse problems and extend them to the blind settings in SISR with significant modifications. To be specific, BlindDiff formulates the blind SR problem under a maximum a posteriori (MAP) framework and unfolds it along with the reverse process in DDPM. Following the MAP optimization paradigm, the key issue now boils down to acquiring the priors of HR clean data and degradation kernels with DDPM. In this case, directly applying conventional pre-trained DDPMs to provide generative priors is not appropriate as they lack the modelling of degradations. Hence, we reconfigure the denoising network and propose the modulated conditional transformer (MCFormer) trained with noise and kernel constraints to guarantee data and kernel priors in each posterior sampling procedure. We then treat the reverse diffusion process as a deep network and introduce a kernel-aware gradient term with respect to the LR condition and two priors, which guide the diffusion model to learn degradation consistency knowledge. Therefore, the denoised images and blur kernels can be jointly optimized by the gradients backpropagated from intermediate states. Besides, such a gradient term can bridge two adjacent iterations to form a closed loop for alternate kernel and HR image estimation until approaching their best approximations with the iteration going on. Moreover, MCFormer injects the modulation between extracted kernel knowledge and image features to exploit their interdependence so that can learn multi-level degradation-aware features for better generation. Attributed to the intrinsic incorporation of MAP-based solution, BlindDiff has highly intuitive physical meanings and is consistent with the classical degradation model (Eq.~(\ref{eq0})), achieving excellent adaptation to various complex degradations in real applications.

The main contributions of this work are four-fold: 
\begin{itemize}
    \item We propose BlindDiff, an effective yet efficient diffusion solver for blind SR. BlindDiff explicitly models the degradation kernels in SISR leveraging the power of DMs, which is expected to achieve high-fidelity generation in real scenarios.
    \item BlindDiff unfolds MAP approach in diffusion models, which enables iteratively alternate optimization for blur kernel estimation and HR image restoration along with the reverse process. 
    \item A modulated conditional transformer (MCFormer) is proposed as the denoising network, which provides data and kernel priors for posterior sampling. Besides, MCFormer incorporates the kernel modulation mechanism to adjust image features in accordance with kernel information, facilitating the acquisition of multi-level degradation-aware prior features.
    \item Extensive experiments on both synthetic and real-world datasets show the SOTA performance of our method. Particularly, it surpasses existing DM-based methods by large margins while heavily decreasing the model complexity (see Figure 1). 
\end{itemize}

%% file: sec/2_formatting.tex
\section{Related Works}
\label{sec:related_works}
\subsection{Blind Image Super-Resolution}
Blind image SR mainly assumes the blur kernel is unknown, which is deemed more applicable in practice. Some methods~\cite{kernelgan,manet,s2k} prioritize individually modelling the blur kernel and perform non-blind SR based on the kernel. KernelGAN~\cite{kernelgan} explores the internal cross-scale recurrence property of image patches to estimate image-specific blur kernels. MANet~\cite{manet} focuses on the locality of degradations and presents to estimate spatial variant kernels from tiny LR image patches. Luo~\emph{et al.}~\cite{dcls} reformulate the classical degradation model and utilize a deep constrained least squares on the estimated kernel to generate clean features. There are also some methods that investigate the interactions between kernel estimation and SR reconstruction in a single network. IKC~\cite{ikc} and DAN~\cite{dan} propose to alternately estimate the blur kernel and SR result conditioned on each other in deep unfolding networks. Fu~\emph{et al.}~\cite{kxnet} propose KXNet which also achieves such a alternate optimization based on the physical generation mechanism between blur kernel and HR image. BSRGAN~\cite{bsrgan} and Real-ESRGAN~\cite{realesrgan} construct synthetic training datasets under more complex degradation models to mimic diverse degradations in real images. 

\subsection{Diffusion Models}
Diffusion model (DM) is derived in~\cite{dm}, which aims to reverse the noising schedule iteratively to generate desired samples. From the arising of DDPM~\cite{ddpm}, DMs have garnered widespread acclaim as generative models and achieved impressive capabilities in various image restoration tasks. SRDiff~\cite{srdiff} and SR3~\cite{sr3} utilize conditional DMs to tackle SISR problem, surpassing previous GAN-based baselines~\cite{esrgan}. Rombach~\emph{et al.}~\cite{ldm} propose latent diffusion models (LDM) operating on learned latent space to purse efficient and effective image synthesis. Kawar~\emph{et al.}~\cite{ddrm} firstly employ pre-trained DDPMs as learned generative prior to solve multiple linear inverse problems including SISR. Chung~\emph{et al.}~\cite{mcg} impose a manifold constraint gradient term put together with the score function to ensure the sampling iterations fall into the manifold. This method is further extended to general noisy inverse problems in~\cite{dps} with the approximation of posterior sampling. Apart from these non-bind diffusion solvers, GDP~\cite{gdp} generalizes DDPM to image enhancement and blind restoration using a blind degradation estimation strategy. It still remains a challenge to overcome complicated degradations for blind SR. A closely related method to this work is BlindDPS~\cite{blinddps} for blind image deblurring, which utilizes two pre-trained DDPMs to provide data and kernel priors and constructs parallel reverse paths to simultaneously restore the kernel and image. Unlike BlindDPS, our method only requires one well-trained DM to learn the two priors and enables alternate optimization for kernel estimation and SR reconstruction in a single reverse process. Besides, based on designed MCFormer, our method can save considerable model parameters while ensuring plausible textures.

%% file: sec/3_preliminary.tex
\section{Preliminary}
\textbf{Diffusion models (DM)}~\cite{dm,ddpm}, or called denoising probabilistic diffusion models (DDPM), are families of generative models, which comprise a diffusion process and a reverse process. Given a clean image $\mathbf{x}_0$, the diffusion process defines a Markov chain $\mathbf{x}_{0:T}$ that gradually adds noise to $\mathbf{x}_0\sim q(\mathbf{x}_0)$ with a fixed variance schedule $\beta_1, ...,\beta_T$
\begin{equation}
q(\mathbf{x}_t|\mathbf{x}_{t-1})=\mathcal{N}(\mathbf{x}_t;\sqrt{1-\beta_t}\mathbf{x}_{t-1},\beta_t\mathbf{I})
\label{eq2}
\end{equation}
where $\mathbf{x}_T \in  \mathcal{N}(0,\mathbf{I})$. Thanks to the property of Markov chain, we can directly acquire $\mathbf{x}_t$ from $\mathbf{x}_0$ with one forward process $q(\mathbf{x}_t|\mathbf{x}_0)=\mathcal{N}(\mathbf{x}_t;\sqrt{\bar{\alpha}_t}\mathbf{x}_0,(1-\bar{\alpha}_t)\mathbf{I})$, where $\alpha_t=1-\beta_t$ and $\bar{\alpha}_t=\prod_{i=0}^{t}\alpha_i$. 

In the reverse process, there is an opposite parameterized Markov chain $\mathbf{x}_{T:0}$, which starts with a random Gaussian noise $\mathbf{x}_T\in  \mathcal{N}(0,\mathbf{I})$ and gradually samples from noised data until reaching $\mathbf{x}_0$
\begin{equation}
		p_\theta(\mathbf{x}_{t-1}|\mathbf{x}_t)=\mathcal{N}(\mathbf{x}_{t-1};\mu_\theta(\mathbf{x}_t,t),\sigma^2_{\theta}(\mathbf{x}_t,t)\mathbf{I})
\label{eq3}
\end{equation}
where $\sigma^2_{\theta}(\mathbf{x}_t, t)=\frac{1-\bar{\alpha}_{t-1}}{1-\bar{\alpha}_t}\beta_t$ denotes the variance set to a time-dependent constant. The mean $\mu_\theta(\mathbf{x}_t, t)=\frac{1}{\sqrt{\alpha_t}}(\mathbf{x}_t-\frac{\beta_t}{\sqrt{1-\bar{\alpha}_t}}\epsilon_\theta(\mathbf{x}_t,t))$ is a learnable parameter estimated by a $\theta$-parameterized denoising network $\epsilon_\theta(\mathbf{x}_t,t)$. Practically, we usually first predict $\tilde{\mathbf{x}}_0$ from $\mathbf{x}_t$, and sample $\mathbf{x}_{t-1}$ using both $\mathbf{x}_t$ and $\tilde{\mathbf{x}}_0$ computed as 
\begin{equation}
\begin{aligned}
\tilde{\mathbf{x}}_0 =\frac{1}{\sqrt{\alpha_t}}(\mathbf{x}_t - \sqrt{1-\bar{\alpha}_t}\epsilon_{\theta}(\mathbf{x}_t,t) \\
q(\mathbf{x}_{t-1}) = \mathcal{N}(\mathbf{x}_{t-1};\mu_t(\mathbf{x}_t, \tilde{\mathbf{x}}_0), \sigma^2\mathbf{I})
\end{aligned}
\label{eq4}
\end{equation}
where the model is trained using the simplified objective function $\mathbb{E}_{t, \mathbf{x}_0,\epsilon}\left[\Vert\epsilon-\epsilon_{\theta}(\sqrt{\bar{\alpha_t}\mathbf{x}_0}+\sqrt{1-\bar{\alpha}_t}\epsilon,t)\Vert^2_2\right]$. Here, $\epsilon\sim \mathcal{N}(\mathbf{0}, \mathbf{I})$.

\begin{figure*}[t]
\centering
\includegraphics[width=0.92\linewidth]{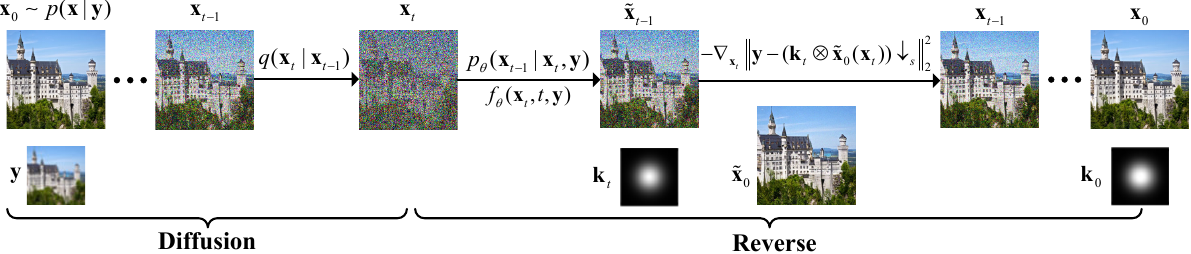}
\caption{Overview of the forward diffusion process (left$\rightarrow$right) that gradually adds Gaussian noise to the original clean image $\mathbf{x}_0$. The diffusion model is trained by $f_{\theta}(\mathbf{x}_t,t,\mathbf{y})$ to provide bot kernel and image priors. In the reverse process, for each timestep $t$, we first sample $\tilde{\mathbf{x}}_{t-1}$ from $\mathbf{x}_t$ and then minimize the residual $\Vert \mathbf{y}-(\mathbf{k}_t\otimes \tilde{\mathbf{x}}_0(\mathbf{x}_t))\downarrow_s\Vert^2_2$ with denoised $\tilde{\mathbf{x}}_{t-1}$, leading to $\mathbf{x}_{t-1}$, where the gradient $\nabla_{\mathbf{x}_t}$ is backpropagated through the whole network. }
\label{fig2}
\end{figure*}

\textbf{Conditional Diffusion Models} aim at generating a target image $\mathbf{x}_0$ from a diffused noise image $\mathbf{x}_T\sim \mathcal{N}(\mathbf{0}, \mathbf{I})$ based on a given condition $\mathbf{y}$, which train a denoising function $\epsilon_{\theta}(\mathbf{x_t,} t, \mathbf{y})$ to learn the conditional distribution $p_{\theta}(\mathbf{x}_t|\mathbf{x}_{t-1},\mathbf{y})$. Once model trained, one can gradually produce a sequential samples $\mathbf{x}_{T:0}$ for the form $\mathbf{x}_0\sim p(\mathbf{x}|\mathbf{y})$
\begin{equation}
\mathbf{x}_{t-1}= \frac{1}{\sqrt{\alpha_t}}(x_t-\frac{\beta_t}{\sqrt{1-\bar{\alpha}_t}}\epsilon_\theta(\mathbf{x}_t,t,\mathbf{y})) + \sqrt{1-\bar{\alpha}_t}\epsilon_t
\label{eq5}
\end{equation}
where $\epsilon_\theta(\mathbf{x}_t,t,\mathbf{y}))$ provides an estimation of the gradient of
the data log-density for Langevin dynamics.

%% file: sec/4_method.tex
In this study, we present BlindDiff, which aims to integrate MAP approach and DDPM for blind SR. BlindDiff decomposes the blind SR problem into two subproblems: \emph{i.e.} kernel estimation and HR image restoration, under a MAP framework. The MAP optimizer is further incorporated into DDPM and unfolded along with the reverse process, achieving alternate estimation for blur kernels and HR images. In this section, we will provide theoretical analysis of the proposed method and introduce BindDiff in detail.

\subsection{Model Formulation}
Given an LR image $\mathbf{y}$ corrupted by the classical degradation model (Eq.~(\ref{eq0})), blind SR involves estimating the blur kernel $\mathbf{k}$ and then reconstructing the HR image $\mathbf{x}$ from $\mathbf{y}$ and $\mathbf{k}$, which can generally formulated as a MAP problem:
\begin{equation}
p(\mathbf{k}, \mathbf{x}|\mathbf{y}) = p(\mathbf{k}|\mathbf{y})p(\mathbf{x}|\mathbf{k}, \mathbf{y})
\label{eq6}
\end{equation}
where $\mathbf{k}$ and $\mathbf{x}$ can be solved by 
\begin{equation}
(\mathbf{k}_0,\mathbf{x}_0)=\mathop{\arg\max}\limits_{\mathbf{k},\mathbf{x}}\log p(\mathbf{k}|\mathbf{y}) + \log p(\mathbf{y}|\mathbf{k}, \mathbf{x})+\log p(\mathbf{x})
\label{eq7}
\end{equation}
where $ \log p(\mathbf{y}|\mathbf{k}, \mathbf{x})$ represents the log-likelihood of $\mathbf{y}$. $\log p(\mathbf{k}|\mathbf{y})$ denotes the kernel prior captured from $\mathbf{y}$ and $\log p(\mathbf{x})$ refers to the clean image prior that is independent of $\mathbf{y}$. $\mathbf{x}_0$ and $\mathbf{k}_0$ denotes the latent optimal estimation of $\mathbf{x}$ and $\mathbf{k}$ respectively. By decomposing Eq.~(\ref{eq7}) into two subproblems, we can get
\begin{subequations}
\begin{numcases}{\centering}
		\mathbf{k}_0=\mathop{\arg\max}\limits_{\mathbf{k}}\log p(\mathbf{k}|\mathbf{y})\label{eq8a}\\
		\mathbf{x}_0=\mathop{\arg\max}\limits_{\mathbf{x}}\log p(\mathbf{y}|\mathbf{k}_0, \mathbf{x}) + \log p(\mathbf{x})\label{eq8b}
\end{numcases}
\end{subequations}
To make a further step, we can unfold Eq.~(8) to form an iterative optimization process
\begin{subequations}
\begin{numcases}{\centering}
		\mathbf{k}_{i-1}=\mathop{\arg\max}\limits_{\mathbf{k}}\log p(\mathbf{k}|\mathbf{y})\label{eq9a}\\
		\mathbf{x}_{i}=\mathop{\arg\max}\limits_{\mathbf{x}}\log p(\mathbf{y}|\mathbf{k}_{i-1}, \mathbf{x}) + \log p(\mathbf{x})\label{eq9b}
\end{numcases}
\end{subequations}

As we can see, in this formula, the estimation for $\mathbf{k}$ is independent to $\mathbf{x}$, thus the overall process can be seen as a two-step blind SR pipeline, \emph{i.e.} kernel estimation from $\mathbf{y}$, and kernel-based non-blind SR. This has been demonstrated to be suboptimal as missing the information from $\mathbf{x}$ for kernel estimation~\cite{dan,dssr}.  To address this issue, we follow the common practice in existing end-to-end blind SR methods that incorporate the latent SR image as an auxiliary information to estimate $\mathbf{k}$. Therefore, Eq.~(\ref{eq8a}) can be reformulated to 
\begin{equation}
\mathbf{k}_{i-1}=\mathop{\arg\max}\limits_{\mathbf{k}}\log p(\mathbf{k}|\mathbf{y}, \mathbf{x}_{i-1})
\label{eq10}
\end{equation}
In addition, since SISR is a conditional image generation task, we often model the condition posterior distribution $p(\mathbf{x}|\mathbf{y})$ as the data prior term rather than $p(\mathbf{x})$. This can be achieved by many well-trained generative models. This work employs DDPMs for that purpose as their excellent generative capabilities.

\subsection{Unfolding MAP Approach in DDPM}
In the DDPM fashion, we can naturally obtain a sequence denoiser prior $(\tilde{\mathbf{x}}_{T-1},...,\tilde{\mathbf{x}}_{t},...,\tilde{\mathbf{x}}_0)$ from $\mathbf{x}_T$ by train a conditional model. Here, let's first analyze the solution for $\mathbf{x}_0$. With the prior $\tilde{\mathbf{x}}_t$ and LR condition $\mathbf{y}$, to solve $\mathbf{x}$, the other key point lies into modelling $\log p(\mathbf{y}|\mathbf{k}, \mathbf{x})$ in the reverse sampling process. 

To achieve this, we firstly specify the posterior distribution of this term to be Gaussian: $p(\mathbf{y}|\mathbf{k}, \mathbf{x})=\mathcal{N}(\mathbf{y}|\mathbf{k}\otimes \mathbf{x}, \sigma^2\mathbf{I})$. According to the diffusion posterior sampling algorithm~\cite{dps} for non-blind inverse problems, we modify it to adapt to blind settings, expressed by
\begin{equation}
\log p(\mathbf{y}|\mathbf{k}_t, \mathbf{x}_t)\approx \nabla_{\mathbf{x}_t}\log p(\mathbf{y}|\mathbf{k}_t,  \tilde{\mathbf{x}}_0(\mathbf{x}_t))
\label{eq11}
\end{equation}
where $\tilde{\mathbf{x}}_0(\mathbf{x}_t)$ is the denoised sample of $\mathbf{x}_t$ computed by Eq.~(\ref{eq4}). $\nabla_{\mathbf{x}_t}\log p(\mathbf{y}|\mathbf{k}_t,  \tilde{\mathbf{x}}_0(\mathbf{x}_t))$ can be obtained through analytical likelihood for our Gaussian measurement model 
\begin{equation}
\nabla_{\mathbf{x}_t}\log p(\mathbf{y}|\mathbf{k}_t,  \tilde{\mathbf{x}}_0(\mathbf{x}_t))=-\frac{1}{\sigma^2}\nabla_{\mathbf{x}_t}\Vert \mathbf{y}-(\mathbf{k}_t\otimes \tilde{\mathbf{x}}_0(\mathbf{x}_t))\downarrow_s\Vert^2_2
\label{eq12}
\end{equation}
Therefore, we can calculate the gradient $\nabla_{\mathbf{x}_t}$ and backpropagate it through the network to optimize for $p(\mathbf{y}|\mathbf{k}_{t}, \tilde{\mathbf{x}}_0(\mathbf{x}_t))$, as shown in Figure~\ref{fig2}. Noted that, if we convert Eq.~(\ref{eq12}) into a solution of minimizing the energy function $\mathbf{x}_i=\mathop{\arg\min}\limits_{\mathbf{x}}\frac{1}{2\delta^2}\Vert \mathbf{y} - (\mathbf{k}_{i-1}\otimes \mathbf{x})\downarrow_s\Vert^2+\phi(\mathbf{x})$, called \textbf{kernel-aware gradient term}, is actually very close to the data fidelity term $\frac{1}{2\delta^2}\Vert \mathbf{y} - \mathbf{k}_{i-1}\otimes \mathbf{x}\Vert^2$ that guarantees the solution to be consistent with the degradation knowledge. 

\begin{figure*}[t]
\centering
\includegraphics[width=0.98\linewidth]{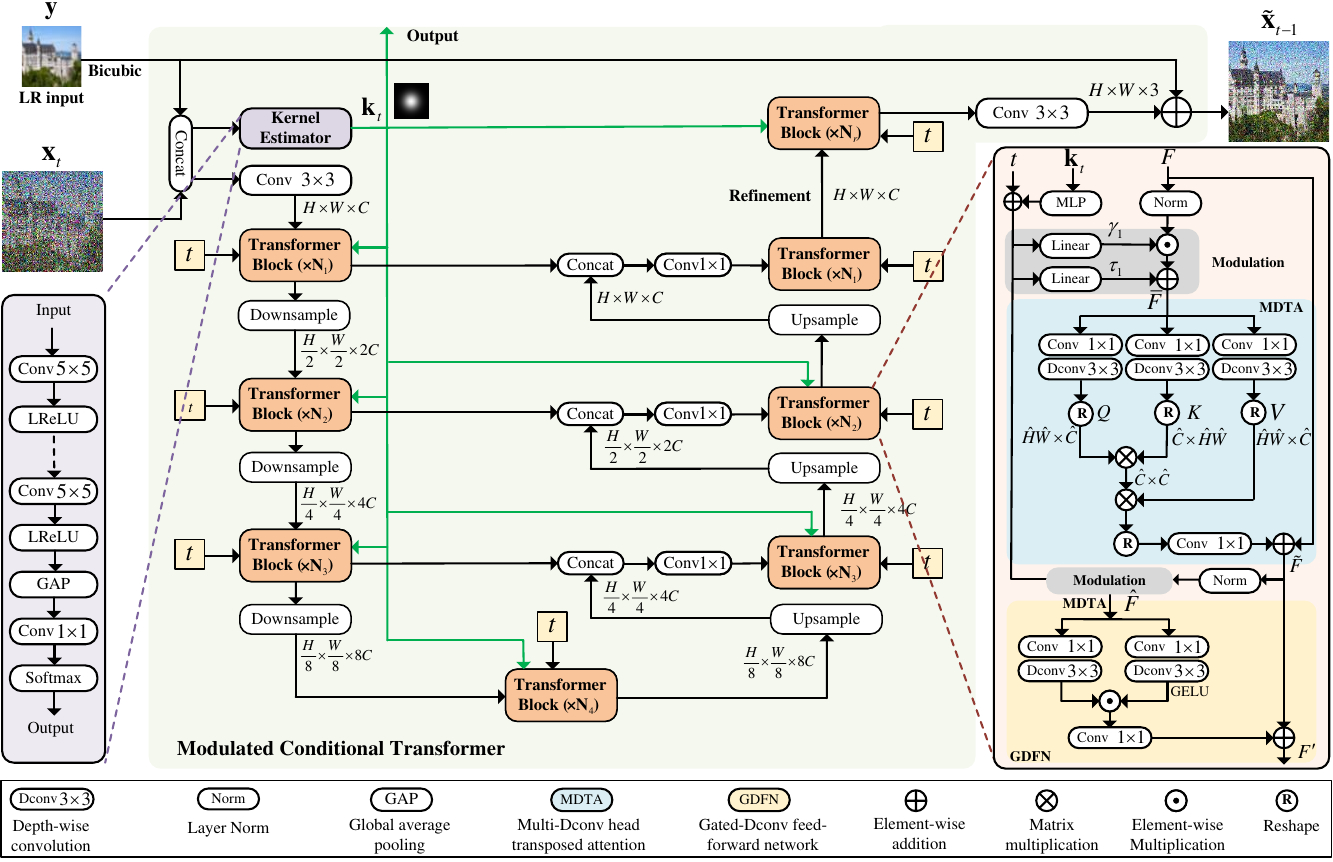}
\caption{Network Architecture of the proposed modulated conditional transformer (MCFormer) in BlindDiff. MCFormer consists of a kernel estimator to predict blur kernels and serial transformer blocks in a multi-scale hierarchical design. }
\label{fig3}
\vspace{-0.2cm}
\end{figure*}

Accordingly, what we need is to provide a data prior $\mathbf{x}_t$ as well as a kernel prior $\mathbf{k}_t$. Combining Eq.~(\ref{eq5}) and Eq.~(\ref{eq9b}), since the estimation for both components relies on $\mathbf{y}$ and previous sample $\mathbf{x}_t$,we propose to train the denoising network with not only the noise-related objective $\epsilon_{\theta}(\mathbf{x}_t, t,\mathbf{y})$ but also a kernel-related objective. To this end, a modulated conditional transformer (MCFormer, see Figure~\ref{fig3}) is proposed consisting a kernel estimator $\mathcal{K}$ and a kernel-modulated transformer backbone, which is trained $\epsilon_{\theta}(\mathbf{x}_t, t,\mathbf{y})$ and the $\mathbf{L}_1$ loss between the ground truth kernel $\mathbf{k}_{gt}$ and the estimated $\mathbf{k}$, formulated by 
%
\begin{equation}
\begin{aligned}
f_{\theta}(\mathbf{x}_0,\mathbf{y},\epsilon)&=\mathbb{E}_{\mathbf{x}_0,\mathbf{y},\epsilon}\Vert \epsilon-\epsilon_{\theta}(\mathbf{y}, \underbrace{\sqrt{\bar{\alpha}}_t\mathbf{x}_0+\sqrt{1-\bar{\alpha}_t}\epsilon}_{\mathbf{\tilde{x}}}, \alpha_t)\Vert^2_2\\
&+\Vert \mathbf{k}_{gt}-\underbrace{\mathcal{K}(\mathbf{y, \sqrt{\bar{\alpha}}_t\mathbf{x}_0+\sqrt{1-\bar{\alpha}_t}\epsilon}}_{\mathbf{k}})\Vert^1_1
\label{eq13}
\end{aligned}
\end{equation}
where $f_{\theta}(\mathbf{x}_0,\mathbf{y},\epsilon)$ denotes the combined total loss and $\tilde{\mathbf{x}}$ is the noisy target image. 

\begin{algorithm}[t]
\renewcommand{\algorithmicrequire}{\textbf{Input:}}
\caption{BlindDiff --- Training a DM $f_{\theta}(\mathbf{x}_0, \mathbf{y}, \epsilon)$}
\begin{algorithmic}[1]
\REQUIRE LR image $\mathbf{y}$ as a condition
\REPEAT
\STATE $(\mathbf{x}_0, \mathbf{y})\sim p(\mathbf{x}, \mathbf{y})$
\STATE $(\mathbf{k}_0, \mathbf{y})\sim p(\mathbf{k}, \mathbf{y}, \mathbf{x})$
\STATE $\epsilon\sim \mathcal{N}(\mathbf{0},\mathbf{I})$
\STATE $T\sim$ Uniform($\{1,...,T\}$)
\STATE Take gradient descent step on $f_{\theta}$
\STATE \quad $\nabla_{\theta}(\Vert \epsilon-\epsilon_{\theta}(\mathbf{y}, \sqrt{\bar{\alpha}}_t\mathbf{x}_0+\sqrt{1-\bar{\alpha}_t}\epsilon, \alpha_t)\Vert^2_2)+ \Vert \mathbf{k}_{gt}-\mathcal{K}(\mathbf{y, \sqrt{\bar{\alpha}}_t\mathbf{x}_0+\sqrt{1-\bar{\alpha}_t}\epsilon})\Vert^1_1$
\UNTIL converged
\end{algorithmic}\label{alg1}
\end{algorithm}
\begin{algorithm}[t]
\renewcommand{\algorithmicrequire}{\textbf{Input:}}
\caption{BlindDiff --- Inference in $T$ iterations}
\begin{algorithmic}[1]
\REQUIRE LR image $\mathbf{y}$ as a condition, a constant $\lambda$
\STATE $(\mathbf{y}|\mathbf{k}, \mathbf{x})\sim \mathcal{N}(\mathbf{y}|\mathbf{k}\otimes \mathbf{x}, \sigma^2\mathbf{I})$
\STATE $\mathbf{x}_T\sim \mathcal{N}(\mathbf{0}, \mathbf{I})$
\FOR {$t=T,...,1$} 
\STATE $\epsilon\sim \mathcal{N}(\mathbf{0}, \mathbf{I})$ if $t>1$, else $\epsilon=\mathbf{0}$
\STATE $\tilde{\mathbf{x}}_{t-1}=\frac{1}{\sqrt{\alpha_t}}(\mathbf{x}_t-\frac{\beta_t}{\sqrt{1-\bar{\alpha}_t}}\epsilon_\theta(\mathbf{x}_t, t, \mathbf{y}))$ \& $\mathbf{k}_t$ generated by $f_{\theta}$
\STATE $\mathbf{x}_{t-1}=\tilde{\mathbf{x}}_{t-1}-\lambda\nabla_{\mathbf{x}_t}\Vert \mathbf{y}-\mathbf{k}_t\otimes \tilde{\mathbf{x}}_0(\mathbf{x}_t)\Vert^2_2$
\ENDFOR
\STATE {\bfseries return} $\mathbf{x}_0, \mathbf{k}_0$
\end{algorithmic}\label{alg2}
\end{algorithm}

Once trained, as illustrated in Figure~\ref{fig2}, at each reverse iteration, the model enables denoised HR image generation and kernel estimation. Then, by bridging the adjacent iterations with the gradient term $\nabla_{\mathbf{x}_t}\log p(\mathbf{y}|\mathbf{k}_t,  \tilde{\mathbf{x}}_0(\mathbf{x}_t))$ (Eq.~(\ref{eq12})), we can form a close loop that estimats the blur kernel and HR images in a mutual learning manner. With the iteration going on, our BlindDiff enables alternate optimization for the two until approaching their best approximations. Correspondingly, Eq.~(9) can be further rewritten as follows
\begin{subequations}
\begin{numcases}{\centering}
		\mathbf{k}_{t}=\mathop{\arg\max}\limits_{\mathbf{k}}\log p(\mathbf{k}|\mathbf{y}, \mathbf{x}_{t})\label{eq14a}\\
		\mathbf{x}_{t-1}=\mathop{\arg\max}\limits_{\mathbf{x}}\log p(\mathbf{y}|\mathbf{k}_{t}, \tilde{\mathbf{x}}_0(\mathbf{x}_t)) + \log p(\mathbf{x}_t|\mathbf{y})\label{eq14b}
\end{numcases}
\end{subequations}

In summary, BlindDiff unfolds the MAP approach into DDPMs, which decouples the blind SR problems into two subproblems that can be alternate solved within the reverse process. The algorithm pseudo-codes of the diffusion and reverse processes in BlindDiff are summarized in Algorithm~\ref{alg1} and Algorithm~\ref{alg2}. Notably, this work is quite different to BlindDPS~\cite{blinddps} that holds $\mathbf{x}_t$, $\mathbf{k}_t$ are independent and conducts parallel reverse using two independent pre-trained DDPMs to provide kernel and data priors. In Table~\ref{tab:blur} and Figure~\ref{fig:blur}, we have demonstrate the superiority of our BlindDiff on the image deblurring task.

\subsection{Modulated Conditional Transformer}
The detailed architecture of our denoising network, \emph{i.e.} modulated conditional transformer (MCFormer), is shown in Figure~\ref{fig3}. MCFormer employs a kernel estimator to learn blur kernels and a multi-scale hierarchical backbone incorporating transformer blocks to model $f_{\theta}(\mathbf{x}_0, \mathbf{y}, \epsilon)$ .

\textbf{Kernel Estimator}. Given an LR input $\mathbf{y}$, we first upsample it to the target resolution using bicubic interpolation and concatenate it with the intermediate image $\mathbf{x}_t$ together. The kernel estimator $\mathcal{K}$ is placed at the beginning of MCFormer for the modulation in later stages. Specifically, it goes through four $5\times 5$ convolutional layers with Leaky ReLU activations (LReLU) to encode image features. Then, we conduct global average pooling on the feature maps to form the estimation elements of the blur kernel, in which the dimensionality of the kernel space is further reduced by principle component analysis (PCA) following~\cite{srmd,ikc}. The whole estimation process is formulated as
\begin{equation}
\mathbf{k}_t=\mathcal{K}(\mathbf{y}{\uparrow_s}, \mathbf{x}_t)
\label{eq19}
\end{equation}
where $\uparrow_s$ denotes bicubic upsamling with scale factor $s$ and $\mathbf{k}_t$ is the resulted kernel at timestep $t$.

\textbf{Kernel-Modulated Transformer Block}. MCFormer embeds the kernel and noise time information into transformer blocks to learn degradation-aware feature presentations. There are three components in the transformer block: 1) Modulations for both the 2) Multi-Dconv head Transposed Attention (MDTA) layer and 3) Gated-Dconv Feed-forward Network (GDFN) layer, as sketched in Figure~\ref{fig3} (right). Let $F\in\mathbb{R}^{H\times W\times C}$ be the intermediate input tensor, where $H$, $W$, and $C$ denotes the height, width, and channel dimension of $F$ respectively. The time $t$ and $\mathbf{k}_t$ are another two inputs. It is notable that $t\in\mathbb{R}^C$ is already pre-processed by a multi-layer perceptron (MLP) before embedding. Then we leverage another MLP to project $\mathbf{k}_t$ to the same dimension as $t$. The first modulation module provides affine transformation for $F$ after layer normalization
\begin{equation}
\bar{F}=\mathcal{M}_1(t,\mathbf{k}_t,F)=\gamma_1 \odot Norm(F) + \tau_1
\end{equation}
where $\mathcal{M}_1$ denotes the transformation function. $\gamma_1$ and $\tau_1$ are learnable parameters by two linear layers after fusing $t$ and $\mathbf{k}_t$, which serve as a scaling and shifting operations respectively. $\bar{F}$ is the modulated feature. Then, we apply MDTA on $\bar{F}$ for global connectivity modelling as it has been demonstrated to be an efficient yet effective attention mechanism~\cite{restormer}, produced $\tilde{F}\in\mathbb{R}^{H\times W\times C}$. 

After that, another modulation module is adopted to operate on the attentive feature
\begin{equation}
\hat{F} = \mathcal{M}_2(t, \mathbf{k}_t, \tilde{F})=\gamma_2 \odot \tilde{F} + \tau_2
\end{equation}
where $\gamma_2$ and $\tau_2$ also learned from the fused feature of $t$ and $\mathbf{k}_t$. A GDFN~\cite{restormer} is finally utilized to aggregate local features and mine meaningful information from spatially neighbouring pixel positions, resulting output feature $F'\in\mathbb{R}^{H\times W\times}$ which is expected to be more discriminative and robust to various degradations.

%% file: sec/5_experiments.tex
\section{Experiments}

\subsection{Implementation Details}
In our experiments, we train BlindDiff using Adam optimizer~\cite{adam} with mini-batch size 8 and HR patch size $256\times 256$ for 500K iterations. The initial learning rate is set to $2e-4$ and decreases by half every 100K iterations. The proposed MCFormer is a 4-level encoder-coder. From level 1 to level 4, the number of transformer blocks are [2, 3, 6, 8] with the channel number of [48, 96,192, 384], where the attention heads in MDTA are set to [1, 2, 4, 8]. For the refinement stage, there are 2 transformer blocks with 48 channels and single-head MDTAs.

\begin{table*}[t]
\small
\centering
\caption{$4\times$ SR quantitative comparison on datasets with \emph{Gaussian8} kernels. \textbf{Bold}: Best, \underline{underline}: second best.}
\setlength{\tabcolsep}{1.05mm}{\begin{tabular}{c|cc|cc|ccc|ccc|ccc}
\hline
\multirow{2}{*}{\textbf{Method}} & \multicolumn{2}{c|}{\textbf{Set5}} & \multicolumn{2}{c|}{\textbf{BSD100}} & \multicolumn{3}{c|}{\textbf{DIV2K100}} & \multicolumn{3}{c|}{\textbf{FFHQ}} & \multicolumn{3}{c}{\textbf{ImageNet-1K}} \\
\cline{2-14}
~ & {LPIPS$\downarrow$} & {PSNR$\uparrow$} & {LPIPS$\downarrow$} & {PSNR$\uparrow$} & {LPIPS$\downarrow$} & {FID$\downarrow$} & {PSNR$\uparrow$} & {LPIPS$\downarrow$} & {FID$\downarrow$} & {PSNR$\uparrow$} & {LPIPS$\downarrow$} & {FID$\downarrow$} & {PSNR$\uparrow$}\\
\hline
IKC~\cite{ikc} & 0.2459 & 31.67 & 0.3699 & 27.37 & 0.3047 & 64.44 & 28.45 & 0.2443 & 51.73 & 32.01 & 0.2702 & 48.62 & 29.06\\
DAN~\cite{dan} & 0.2321 & 31.90 & 0.3603 & 27.51 & 0.2933 & 59.76 & \underline{28.91} & 0.2167 & 45.33 &  \underline{32.80} & 0.2615 & \underline{47.81} & \underline{29.54}\\
AdaTarget~\cite{adatarget} & 0.2351 & 31.58 &  0.3598 & 27.43 & 0.2919 & 57.72 & 28.80 & 0.2362 & 48.19 & 32.24 & 0.2631 & 56.81 & 29.40\\
BSRGAN~\cite{bsrgan} & 0.3175 & 27.50 & 0.3752 & 25.32 & 0.3371 & 85.47 & 25.47 & 0.2435 & 40.22 & 28.80 & 0.3390 & 73.20 & 26.16\\
DCLS~\cite{dcls} & \underline{0.2284} & \textbf{32.12} & \underline{0.3574} & \underline{27.60} & \underline{0.2855} & \underline{57.57} & \textbf{29.04} & \underline{0.2147} & 45.02 & \textbf{32.86} & 0.2580 & 47.83 & \textbf{29.65}\\
KDSR~\cite{kdsr} & 0.2440 & \underline{32.11} & 0.3759 & \textbf{27.64} & 0.3125 & 72.34 & 28.76 & 0.2543 & 53.10 & 32.29 & 0.2801 & 53.15 & 29.43\\
\rowcolor{gray!30}DiffPIR~\cite{diffpir} & 0.3398 & 26.51 & 0.4470 & 24.96 & 0.4131 & 94.00 & 25.11 & 0.2795 & 44.81 & 28.08 & 0.3747 & 62.70 & 25.77\\
\rowcolor{gray!30}GT kernel+DPS~\cite{dps} & \multicolumn{2}{c|}{---} & \multicolumn{2}{c|}{---} & \multicolumn{3}{c|}{---} & 0.3059 & \underline{33.77} & 24.69 & 0.4521 & 64.87 & 22.13\\
\rowcolor{gray!30} GDP~\cite{gdp} & 0.4089  & 25.15 & 0.5132 & 23.94 & 0.4765 & 80.41 & 23.88 &  0.3731 & 80.39 & 26.70 & \underline{0.4331} & 82.85 & 24.49\\
\rowcolor{gray!30}\textbf{BlindDiff (ours)} & \textbf{0.2147} & 30.17 & \textbf{0.3099} & 26.34 & \textbf{0.2588} & \textbf{34.15} & 26.83 & \textbf{0.1754} & \textbf{22.02} & 30.87  & \textbf{0.2411} &\textbf{37.18} & 28.12 \\
\hline
\end{tabular}}
\label{tab1}
\end{table*}
\begin{figure*}[t]
\centering
\includegraphics[width=1\linewidth]{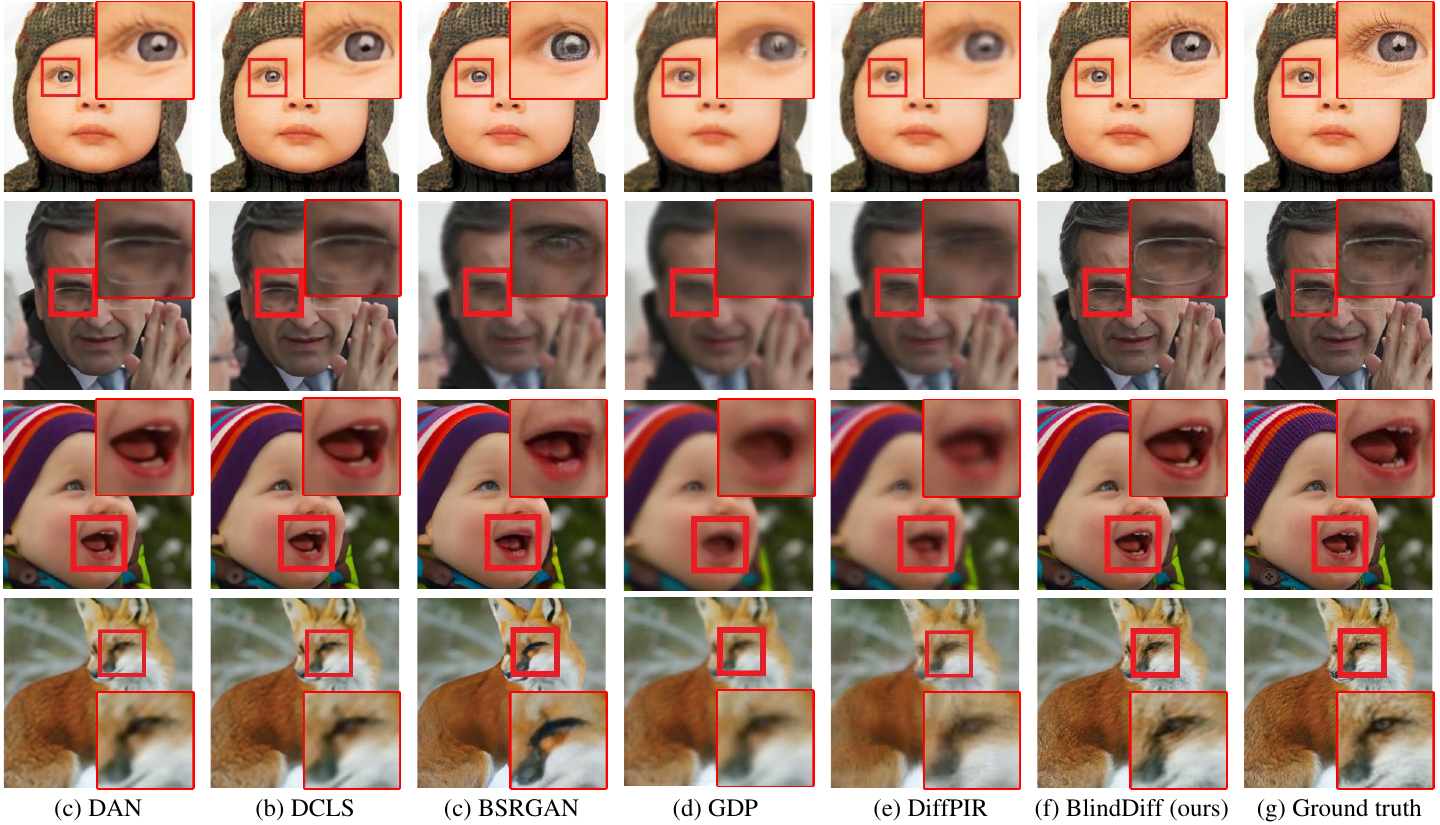}
\caption{Visual comparisons of $4\times$ blind SR methods on isotropic Gaussian kernels.}
\label{fig4}
\vspace{-0.2cm}
\end{figure*}

\subsection{Datasets and Settings}
We apply BlindDiff to the blind SR task and evaluate its $4\times$ SR performance on one face dataset FFHQ $256\times 256$~\cite{ffhq} and six natural image datasets including Set5~\cite{set5}, BSD100~\cite{bsd100}, DIV2K100~\cite{div2k}, and ImageNet-1K $256\times 256$~\cite{imagenet}. FFHQ contains 70000 human face images, from which we randomly select 1000 images for validation and the rest for training. For natural image SR, we train models on DIV2K (800) and Flickr2K (2650) datasets, totally 3450 images. Notably, for efficient comparison, in DIV2K100, the resolution of LR images is set to $128\times 128$ so that the corresponding HR images are of the size $512\times 512$.  We employ learned perceptual image patch similarity (LPIPS)~\cite{lpips}, Fr{\'e}chet inception distance (FID)~\cite{fid} and peak signal-to-noise ratio (PSNR) as the metrics to measure the quantitative performance. In this work, we focus on two cases of degradations: \textbf{1) Isotropic Gaussian blur kernels}. Following~\cite{dan},  in the training phase, we set the blur kernel size to $21\times 21$ and uniformly sample the kernel width in [0.2, 4.0]. During testing, we use the \emph{Gaussian8} kernel set which consists of 8 blur kernels with the range of [1.8, 3.2]; \textbf{2) Anisotropic Gaussian blur kernels}. The blur kernels are set to $11\times 11$ anisotropic gaussians with independently random distribution in the range of [0.6, 5] for each axis and rotated by a random angle in [$-\pi$, $\pi$].  To deviate from a regular Gaussian, the uniform multiplicative noise up to 25\% of each pixel value of the kernel) is then applied and normalized to sum to one.

\begin{figure*}[t]
\centering
\includegraphics[width=1\linewidth]{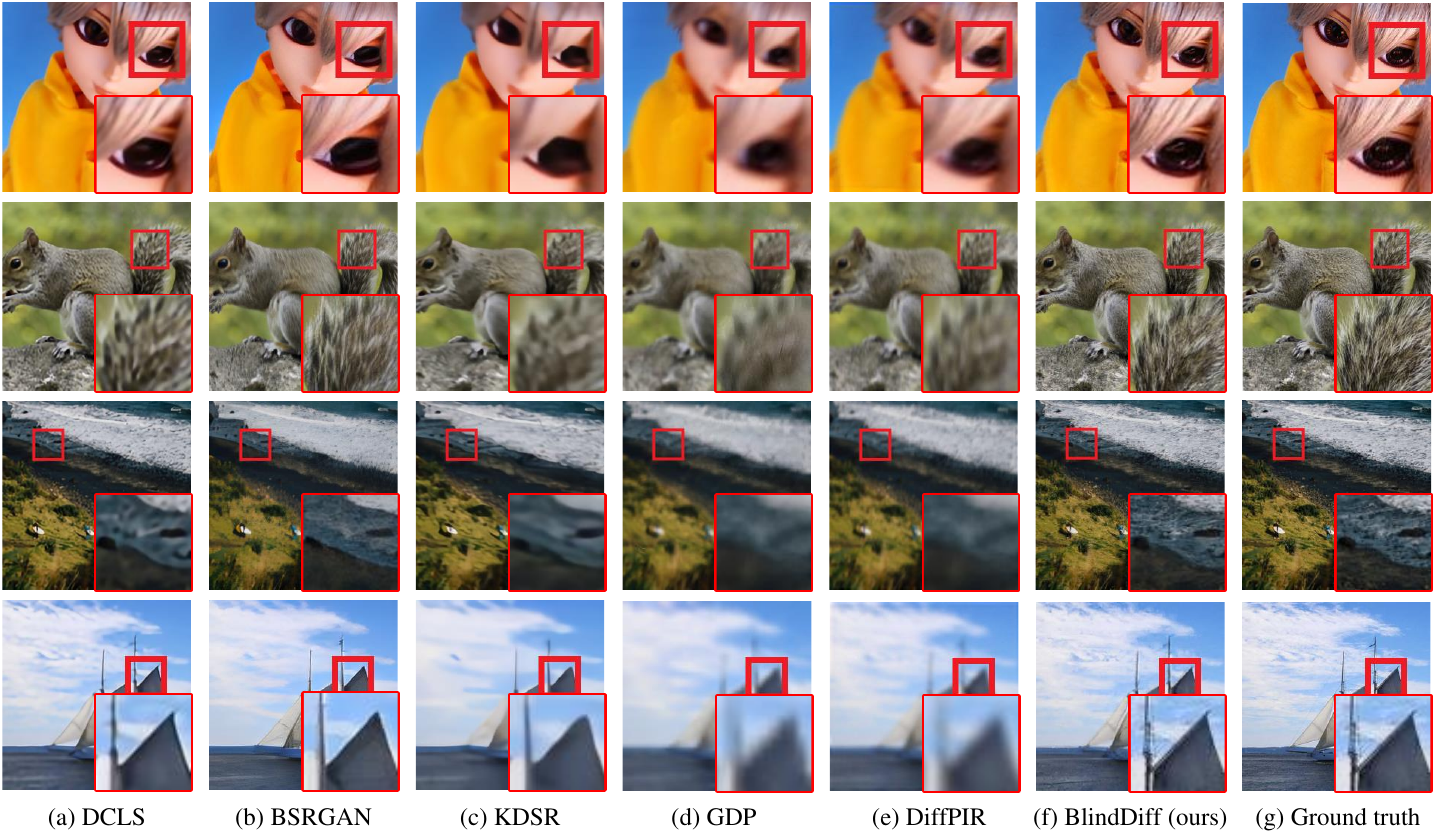}
\caption{Visual comparisons of $4\times$ blind SR methods on anisotropic Gaussian kernels.}
\label{fig5}
\end{figure*}

 \begin{table}[t]
\scriptsize
\centering
\caption{$4\times$ SR quantitative comparison on datasets with random anisotropic Gaussian kernels. \textbf{Bold}: Best, \underline{underline}: second best.
}
\setlength{\tabcolsep}{1.05mm}{\begin{tabular}{c|ccc|ccc}
\hline
\multirow{2}{*}{\textbf{Method}} & \multicolumn{3}{c|}{\textbf{DIV2K100}} & \multicolumn{3}{c}{\textbf{ImageNet-1K}} \\
\cline{2-7}
 ~ & {LPIPS $\downarrow$} & {FID$\downarrow$}  & {PSNR$\uparrow$} & {LPIPS$\downarrow$} & {FID$\downarrow$} & {PSNR$\uparrow$}\\
\hline
DASR~\cite{dasr} & 0.4476 & 149.11 & 25.46 & 0.4116 & 100.66 & 26.22\\
DAN~\cite{dan} & 0.3597 & 96.63 & 26.74 & 0.3272 & 68.52 & 27.33 \\
DCLS~\cite{dcls} & 0.3085 & \underline{69.98} &  \textbf{28.31} & \underline{0.2791} & \underline{54.59} & \textbf{29.02} \\
BSRGAN~\cite{bsrgan} & 0.3526 & 98.39 & 24.90 & 0.3546 & 80.95 & 25.60 \\
AdaTarget~\cite{adatarget} & \underline{0.2923} & 77.04 & \underline{28.25} & 0.3249 & 56.81 & \underline{27.58}\\
DARSR~\cite{darsr} & 0.4956  & 148.34 & 24.05 & 0.4618 & 107.79 & 24.22\\
KDSR~\cite{kdsr} & 0.4328 & 144.25 & 25.82  & 0.4035 & 101.22 & 26.48\\
\rowcolor{gray!30}DiffPIR~\cite{diffpir} & 0.4387 & 113.02 & 24.75 & 0.4049 & 76.18 & 25.31 \\
\rowcolor{gray!30}GT kernel + DPS~\cite{dps} &  \multicolumn{3}{c|}{---} & 0.4550 & 63.47 & 21.99\\
\rowcolor{gray!30} GDP~\cite{gdp}  &  0.4868 & 137.99 & 23.63 & 0.4491 & 90.49 & 24.16 \\
\rowcolor{gray!30}\textbf{BlindDiff~(ours)}  & \textbf{0.2763} & \textbf{42.85} & 26.32 & \textbf{0.2571} & \textbf{42.62} & 27.53 \\
\hline
\end{tabular}}
\label{tab2}
\end{table}

\subsection{Comparing with the State-of-the-Art}
\textbf{Evaluation with Isotropic Gaussian Blur Kernels}. We quantitatively and qualitatively compare BlindDiff with recent state-of-the-art methods including 6 CNN-based blind SR methods: IKC~\cite{ikc}, DAN~\cite{dan}, AdaTarget~\cite{adatarget}, DCLS~\cite{dcls}, BSRGAN~\cite{bsrgan}, KDSR~\cite{kdsr} and 3 DM-based SR methods: DPS~\cite{dps}, GDP~\cite{gdp}, DiffPIR~\cite{diffpir}. All the methods are evaluated on Set5, BSD100, DIV2K100, FFHQ, and ImageNet-1K datasets, which are synthesized by \emph{Gaussian8}. Here, considering the limitation of existing DM methods for tackling arbitrary resolutions (\emph{i.e.} Set5), we apply the patch-based strategy in GDP~\cite{gdp} with adaptive instance normalization to obtain more natural SR results. Due to the heavy computation cost of DPS (out of memory), we only report its performance on FFHQ and ImageNet-1K. As shown in Table~\ref{tab1}, BlindDiff generally outperforms all the methods in terms of FID and LPIPS on all datasets while also surpassing DM methods in terms of all the 3 metrics by large margins. Specifically, BlindDiff exceeds BSRGAN by 0.09 LPIPS and 36 FID on ImageNet-1K even though it trained for complicated degradations with paired training data. DiffPIR suffers from severe performance drops when directly applied to blind degraded images. DPS performs better than DiffPIR when GT kernels are available, but still exhibits worse perceptual quality amd pixel faithfulness than ours. Surprisingly, although GDP adopts a degradation-guidance strategy to tackle blind degradation, the results indicate it is not well applicable to blind SR. Figure~\ref{fig4} visualize the $4\times$ SR results on face and natural images. It is obvious that the SR results of BlindDiff are much clearer and reveal more details.

\begin{figure*}[t]
\centering
\includegraphics[width=1\linewidth]{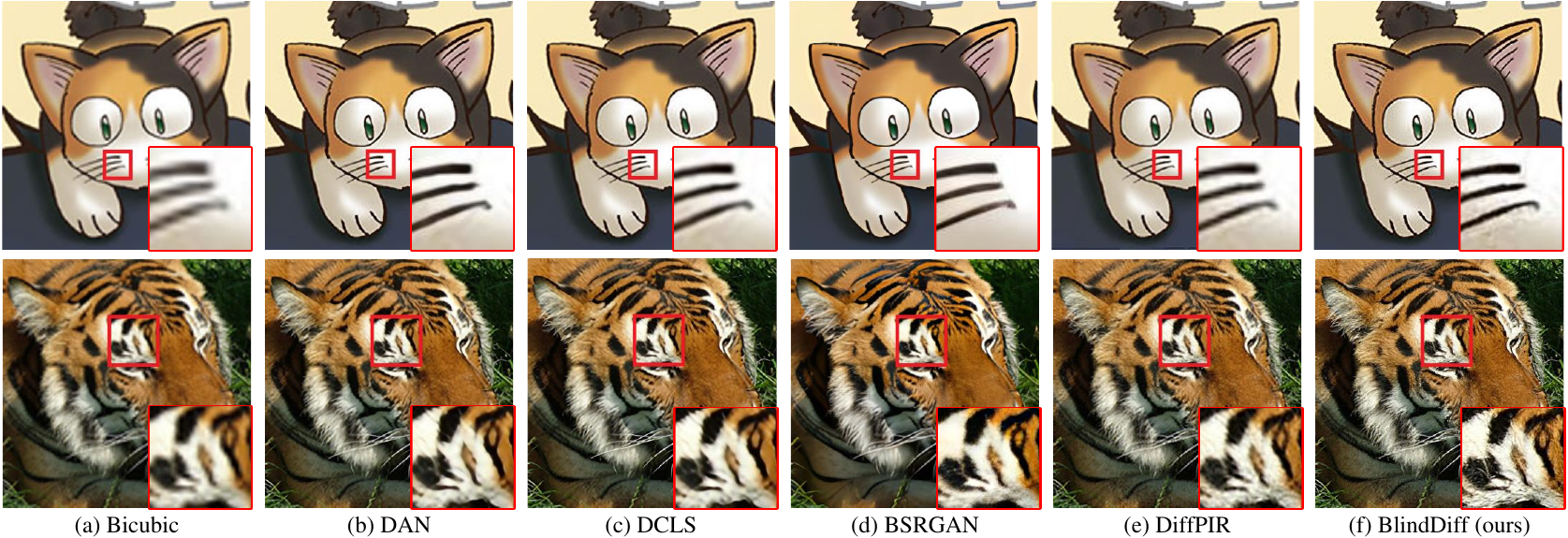}
\caption{Visual results of $4\times$ blind SR methods on RealSRSet. }
\label{fig6}
\end{figure*}

\textbf{Evaluation with Anisotropic Gaussian Blur Kernels}. We evaluate our BlindDiff on DIV2K100 and ImageNet-1K datasets, where the HR images are degraded by random anisotropic Gaussian kernels. We compare BlindDiff with 7 CNN methods: DASR~\cite{dasr}, DAN~\cite{dan}, DCLS~\cite{dcls}, BSRGAN~\cite{bsrgan}, AdaTarget~\cite{adatarget}, DARSR~\cite{darsr}, KDSR~\cite{kdsr} and 3 DM methods: DiffPIR~\cite{diffpir}, DPS~\cite{dps}, GDP~\cite{gdp}. Quantitative results are illustrated in Table~\ref{tab2} and the model complexity measured by parameters and Multi-Adds are shown in Figure~{1}. Clearly, even in this more general and challenging case, BlindDiff still outperforms other methods. Compared to the most SOTA CNN method DCLS, BlindDiff obtains 38.8\% and 21.9\% FID improvements on DIV2K100 and ImageNet-1K respectively. Furthermore, BlindDiff surpasses the best DM method ``GT Kernel+DPS'' by 20.85 in terms of FID while only consuming about 5\% parameters (26.51M \emph{v.s.} 552.81M, Figure 1(a)) and 1\% computations (Figure 1(b)). In addition, the visual comparison of these methods is provided in Figure~\ref{fig5}. As we can see, the SR images of BlindDiff have more realistic textures, fewer blurs, and sharper boundaries than compared methods.

\textbf{Evaluation on Real-World Images}. We also investigate the effectiveness of BlindDiff for real-world degradations by conducting experiments on several real-world images from RealSR~\cite{realsr} and RealSRSet~\cite{bsrgan} datasets, which contains unknown blurs, noise or compression artifacts. The $4\times$ SR results are shown in Figure~\ref{fig6}. We can observe that DAN, DCLS, and DiffPIR tend to generate blurs. Though BSRGAN can effectively remove the noise and blurs, it is easier to produce the SR images with over-smooth textures, leading to details loss. In contrast, our method can recover clearer details and more natural textures.

\begin{table}
\centering
\scriptsize
\caption{Investigation of the Proposed Components in BlindDiff. We evaluate the LPIPS performance on Set5 for $4\times$ blind SR.}
\setlength{\tabcolsep}{1.05mm}{\begin{tabular}{c|c c c c | c}
\hline
\multirow{2}{*}{\textbf{Variant}} & \textbf{Kernel} & \textbf{Kernel} & \multicolumn{2}{c|}{\textbf{Gradient Term}} & \multirow{2}{*}{\textbf{LPIPS$\downarrow$}}\\
\cline{4-5}
~ &  \textbf{Constraint} & \textbf{Modulation} & \textbf{Bicubic-based} & \textbf{Kernel-aware} & ~\\
\hline
Baseline & \XSolidBrush & \XSolidBrush & \XSolidBrush & \XSolidBrush & 0.2382\\
Model A & \CheckmarkBold  & \XSolidBrush &  \CheckmarkBold  & \XSolidBrush & 0.3277\\
Model B & \CheckmarkBold  & \XSolidBrush & \XSolidBrush  &  \CheckmarkBold  & 0.2300\\
\hline
Model C & \CheckmarkBold  & \CheckmarkBold &  \XSolidBrush  & \XSolidBrush  & 0.2298\\
Model D & \CheckmarkBold  & \CheckmarkBold & \XSolidBrush & \CheckmarkBold & \textbf{0.2289}\\
Model E & \CheckmarkBold  & \CheckmarkBold & \CheckmarkBold & \XSolidBrush & 0.3236\\
Model F &  \XSolidBrush & \XSolidBrush & \CheckmarkBold & \XSolidBrush & 0.3268\\
\hline
\end{tabular}}
\label{tab:modulation}
\vspace{-0.4cm}
\end{table}

\subsection{Ablation Study}
In this section, we verify the effectiveness of each component in BlindDiff. For efficient comparison, all the models are pre-trained for 100K iterations and tested on Set5.

\textbf{Effect of Kernel Prior}. In BlindDiff, we propose the MCFormer trained with noise and kernel constraints (Eq.~(\ref{eq13})) to provide additional kernel prior. To investigate its effect, we first construct a \textbf{Baseline} model that uses none of our proposed component. As shown in Table~\ref{tab:modulation}, when we add the kernel prior but use the bicubic-based gradient term (\textbf{Model A}), the model suffers from performance drop due to the difference between the learned kernel and bicubic one. When cooperating with the more related kernel-aware gradient term (\textbf{Model B}), the LPIPS is obviously improved. Continuously, by combining it with the kernel modulation operations, leading to \textbf{Model C}, compared to the \textbf{Baseline}, the result not only proves that effectively utilizing kernel prior knowledge in hidden space is beneficial for better SR reconstruction, but also demonstrates the effectiveness of the modulation operation.

\textbf{Effect of Kernel-aware Gradient term}. The kernel-aware gradient term is the key to achieve alternate kernel and HR image optimization.  Firstly, the comparison between \textbf{Model A} and \textbf{Model B} indicates its superiority for blind blur kernels against the bicubic-based. When we combine it with the kernel constraint and modulation in one model (\textbf{Model D}), our BlindDiff can fully exploit the kernel knowledge so that achieves the best LPIPS. To more intuitively illustrate the influence of this component, we replace it with bicubic-based term in \textbf{Model D}, resulting in \textbf{Model E}, one can see the LPIPS value is heavily decreased. More extremely, we remove all the components but only add the bicubic-based term (\textbf{Model F}), the result further indicates the adverse effects of the kernel difference and the necessity of our kernel-aware gradient term.

\begin{figure*}
  \centering
   \subfloat[]{
    \label{fig7:a}
    \includegraphics[width=0.475\linewidth]{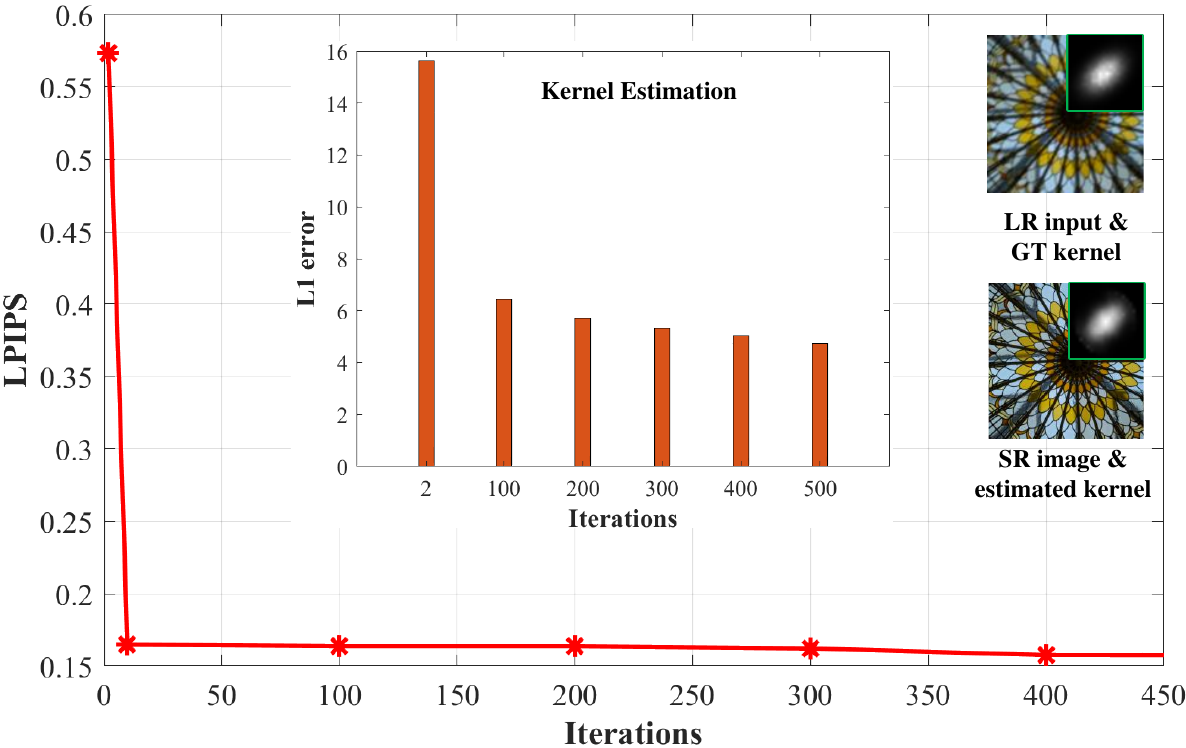}
}
   \subfloat[]{
   \label{fig7:b}
    \includegraphics[width=0.49\linewidth]{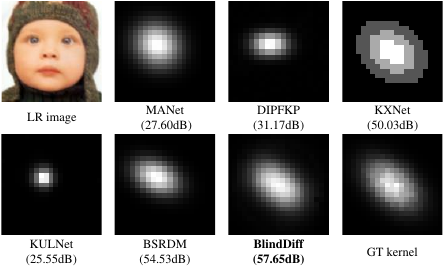}
}
  \caption{(a) Visualization of the results under different sampling iterations. We report the LPIPS and L1 error to measure the estimation accuracy of the SR images and blur kernels, respectively; (b) Kernel estimation results on ``\emph{baby}'' in Set5 degraded by an anisotropic kernel with $\sigma_x=2.5$ and $\sigma_y=4.0$.}
\end{figure*}

\textbf{Balance for Kernel-Aware Gradient Term}. As shown in Algorithm~\ref{alg2}, in the reverse process, we practically calculate the kernel-aware gradient and backpropagate it with a balance hyperweight $\lambda$. Here, we investigate the impact of this term by change the value of $\lambda$ from 0 to 10, where $\lambda=0$ means the model without considering such a gradient backpropagation strategy. The detailed experimental results are listed in Table~\ref{tab:lambda}. One can see when we increase $\lambda$ from 0 to 1, the gradient term becomes more important and provides positive impact for better SR image reconstruction. However, when we further increase $\lambda$ to 10, the model suffers from performance degradation due to the over-emphasizing of the gradient term. These results validate the effectiveness of the proposed kernel-aware gradient term. In our experiments, we accordingly set $\lambda=1$. 

\begin{table}
\centering
\caption{Impact of the kernel-aware gradient term. We evaluate the LPIPS performance on Set5 for $4\times$ blind SR.}
\begin{tabular}{c|c c c c c}
\hline
$\lambda$ & 0 & 0.1 & 1 & 5 & 10\\
\hline
LPIPS & 0.2298 & 0.2296 & \textbf{0.2289} & 0.2716 & 0.2899\\
\hline
\end{tabular}
\label{tab:lambda}
\vspace{-0.4cm}
\end{table}

\textbf{Sampling Step Visualization}. Since our BlindDiff involves the alternate optimization for blur kernel and HR image during posterior sampling, here, we visualize the intermediate sampling steps to investigate how the iteration number affects BlindDiff. Here, we calculate the LPIPS and L1 error to measure the estimation accuracy of the SR images and blur kernels at different iterations, where the results are visualized in Figure~\ref{fig7:a}. One can see that, as the iteration number is larger than 2, BlindDiff almost keep stable and progressivly reaches the upper bound. Besides, we observe that the L1 errors between the predicted kernel and groud truth (GT) kernel also decreases as the iteration goes on, getting closer to the GT kernel. This visualization of the sampling process validates that our BlindDiff can achieve mutually alternate optimization for blur kernel estimation and HR image restoration.

\textbf{Kernel Estimation}. To reveal the kernel estimation ability of BlindDiff, following~\cite{manet}, we calculate the LR image PSNR to quantitatively compare the kernel loss. Here, we synthesize a reference LR image using a GT kernel and a HR image, where an accurate kernel is expected to reconstruct a similar LR image from the same HR one. In Figure~\ref{fig7:b}, we compare BlindDiff with existing kernel estimation-based blind SR methods including~MANet~\cite{manet}, DIPFKP~\cite{fkp}, KXNet~\cite{kxnet}, KULNet~\cite{kulnet}, and BSRDM~\cite{bsrdm}. One can see that BlindDiff generates a blur kernel that is more consistent with the GT kernel while other methods produce unreliable results. Besides, equipped with the resulted kernel, BlindDiff can reconstruct a LR image with the best PSNR performance.

\begin{figure}
\centering
\includegraphics[width=1\columnwidth]{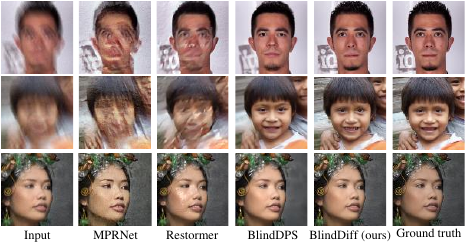}
\caption{Visual comparison for blind deblurring. }
\label{fig:blur}
\end{figure}

\textbf{Extension to Blind Image Deblurring}. To investigate the generalization ability of our method, we conduct experiments on blind deblurring and compared it
with several recent state-of-the-art deblurring methods including MPRNet~\cite{mprnet}, Restormer~\cite{restormer}, and BlindDPS~\cite{blinddps}. Following BlindDPS, motion blur kernel is randomly generated with intensity 0.5 using the code~\footnote{https://github.com/LeviBorodenko/motionblur}. All the methods are evaluated on FFHQ datasets,
where the quantitative and qualitative results are shown in Table~\ref{tab:blur} and Figure~\ref{fig:blur}. As we can see, our BlindDiff not only achieves the best pixel perceptual quality (LPIPS/FID) and faithfulness (PSNR), but also restore the images with more plausible textures.

\begin{table}
\centering
\begin{tabular}{c|ccc}
\hline
\multirow{2}{*}{\textbf{Method}} & \multicolumn{3}{c}{\textbf{FFHQ} ($\bf 256 \times 256$)} \\
\cline{2-4}
& {LPIPS $\downarrow$} & {FID $\downarrow$} & {PSNR $\uparrow$}\\
\hline
MPRNet~\cite{mprnet} & 0.5706 & 185.58 & 21.12\\
Restormer~\cite{restormer} & 0.5050 & 160 87 & 22.36 \\
\rowcolor{gray!30} BlindDPS~\cite{blinddps} & \underline{0.2842} & \underline{30.06} & \underline{24.03} \\
\rowcolor{gray!30}\textbf{BlindDiff~(ours)} & \textbf{0.2542} & \textbf{29.90} & \textbf{26.80} \\
\hline
\end{tabular}
\caption{
Quantitative comparison for blind deblurring on FFHQ for blind deblurring. \textbf{Bold}: Best, \underline{underline}: second best.
}
\label{tab:blur}
\end{table}

%% file: sec/6_conclusion.tex
\section{Conclusion}
In this work, we propose BlindDiff, a DM-based super-resolver that integrates MAP approach into DMs seamlessly to tackle the blind SR problem. Our BlindDiff constructs a distinctive reverse pipeline that unfolds the MAP approach along with the reverse process, enabling alternate optimization for joint blur kernel estimation and HR image recovery. We theoretically analyze the methodology of such a MAP-driven DDPM for blind SR. We present a modulated conditional transformer and make it allows to provide generative kernel and image priors by introducing an kernel estimation objective to train with the noise prediction objective together.  With extensive experiments, we show that BlindDiff establishes state-of-the-art performance.